\pdfoutput=1
\def\year{2022}\relax
\documentclass[letterpaper]{article} 
\usepackage{aaai22}  
\usepackage{times}  
\usepackage{helvet}  
\usepackage{courier}  
\usepackage[hyphens]{url}  
\usepackage{graphicx} 
\usepackage{natbib}  
\usepackage{caption} 
\DeclareCaptionStyle{ruled}{labelfont=normalfont,labelsep=colon,strut=off} 
\usepackage{algorithm}
\usepackage{algorithmic}

\usepackage{newfloat}
\usepackage{listings}
\floatstyle{ruled}
\newfloat{listing}{tb}{lst}{}
\floatname{listing}{Listing}
\title{Scale-aware Two-stage High Dynamic Range Imaging}
\author{Hui Li, Xuyang Yao, Wuyuan Xie, Miaohui Wang}
\usepackage{bibentry}

\begin{document}

\maketitle

\begin{abstract}
Deep high dynamic range (HDR) imaging as an image translation issue has achieved great performance without explicit optical flow alignment. However, challenges remain over content association ambiguities especially caused  by saturation and large-scale movements. To address the ghosting issue and enhance the details in saturated regions, we propose a scale-aware two-stage high dynamic range imaging framework (STHDR) to generate high-quality ghost-free HDR image. The scale-aware technique and two-stage fusion strategy can 
progressively and effectively improve the HDR composition performance.
Specifically, our framework consists of feature alignment and two-stage fusion. 
In feature alignment,  we propose a spatial correct module (SCM) to better exploit useful information among non-aligned features to avoid ghosting  and saturation.
In the first stage of feature fusion, we obtain a preliminary fusion result with little ghosting. In the second stage, we conflate the results of the first stage with aligned features to further reduce residual artifacts and thus improve the overall quality.  Extensive experimental results on the typical test dataset validate the effectiveness of the proposed  STHDR in terms of speed and quality.
\end{abstract}

\section{Introduction}
The dynamic range of natural scenes is much higher than what ordinary digital cameras can capture in one single shot. 
Therefore, overexposure or underexposure often occurs in the photographic experience, resulting in poor imaging results and serious information loss. Extensive attempts have been made to solve the problem of limited dynamic range in both hardware improvement and advanced algorithm design. 
Due to the expensive hardware implementation of HDR imaging, computational photography techniques are commonly used to generate HDR images in mobile terminals. A typical approach is to take multiple exposure sequences and then fuse them to produce a high dynamic range imaging result.

\begin{figure}
	\begin{center}
	    \includegraphics[scale=0.32]{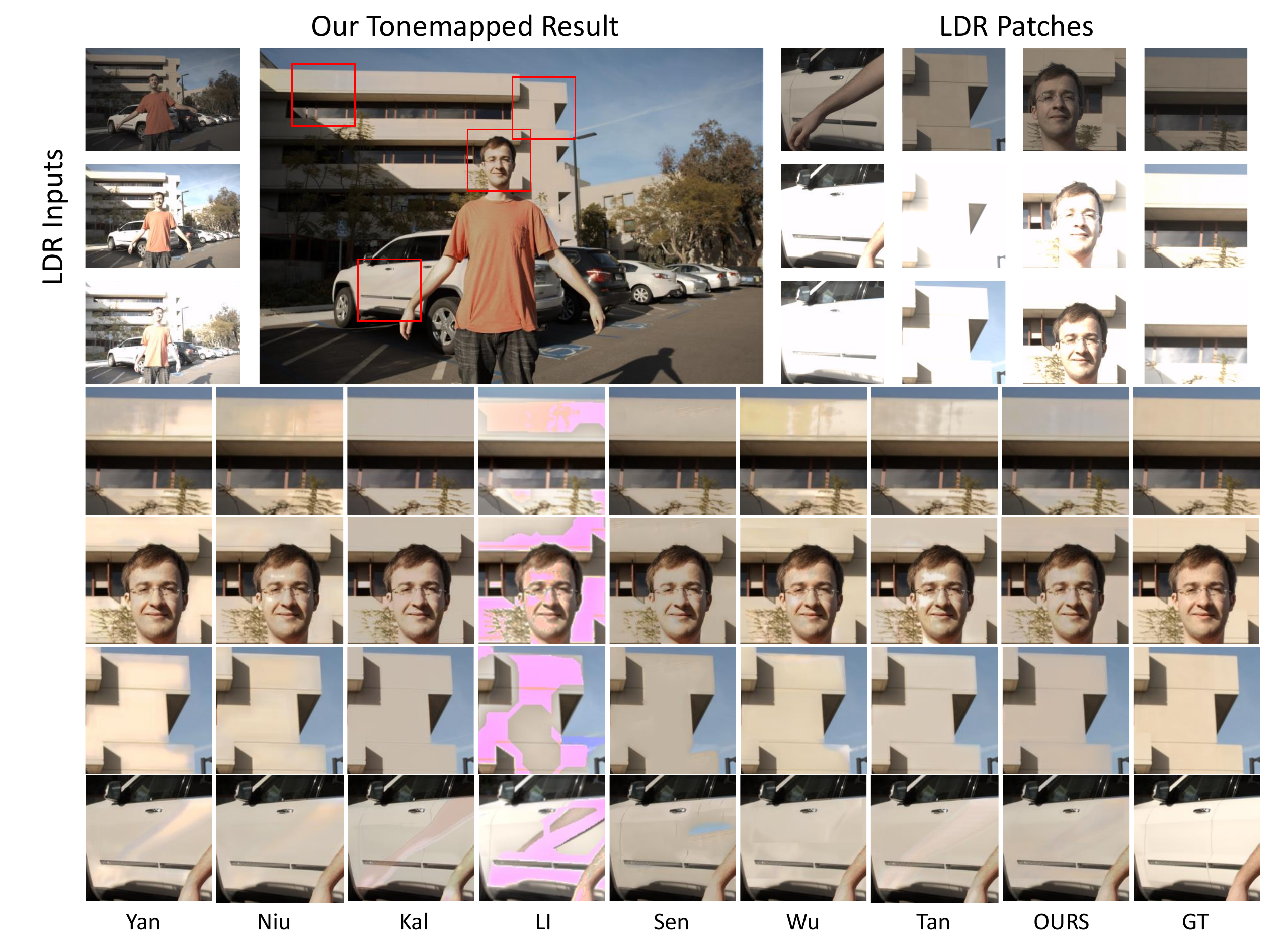}
		\caption{Visual quality comparison of HDR synthesis results via conventional (\cite{li2020fast}, \cite{sen2012robust}) and deep-learning based methods (\cite{yan2019attention}, \cite{niu2021hdr}, \cite{kalantari2017deep}, \cite{wu2018deep}, \cite{tan2021deep}). Deep-learning based methods have less artifacts and more details in saturated regions than conventional methods. The proposed network tends to generate HDR results with more consistency than other deep-learning based methods. 
	}
		\label{fig:example}
	\end{center}
\end{figure}

With multi-exposure images, there are generally two strategies to obtain HDR-like images: Multiple Exposure Fusion (MEF) in the image domain, and intra-HDR reconstruction (CRF) via the camera response function followed by tonal reproduction in the radiometric domain.
Due to the nonlinearity of the image domain, algorithms that reject motion pixels in misaligned regions in the image domain are often prone to artifacts such as ghosting as shown in Figure.~\ref{fig:example}. Therefore, more attention has been paid to achieving HDR imaging in the linear domain. With the success of deep learning on low-level problems, State-of-the-art high dynamic range imaging results are achieved in the linear domain through deep learning.
In \cite{kalantari2017deep}, they were the first to use deep learning to merge LDR inputs. They were prone to generate ghosting artifacts due to unreliable optical flow. ~\cite{wu2018deep} removed optical flow registration and directly learn the transform relationship via an end-to-end network.  Subsequent work about high dynamic range imaging based on deep learning is mainly to continue Wu's~\cite{wu2018deep} method by designing better networks to achieve better high dynamic range imaging results. However, a common problem of these deep HDR imaging is the introduction of ghosting artifacts and detail loss in saturated regions (see Figure.~\ref{fig:example}).
 
To tackle these challenges in deep HDR, we propose a scale-aware two-stage high dynamic range imaging framework in this work. Our main contributions can be summarized as:
\begin{itemize}
\item We propose an implicit feature alignment method called Spatial Correct Module (SCM) that can robustly leverage the useful information in the presence of large movement and saturation. 
\item Two-stage fusion strategy are used to progressively improve the fusion performance.
\item Multi-scale implementation are employed to gain both coarse and fine details.
\end{itemize}

\begin{figure*}
	\begin{center}
		\includegraphics[scale=0.58]{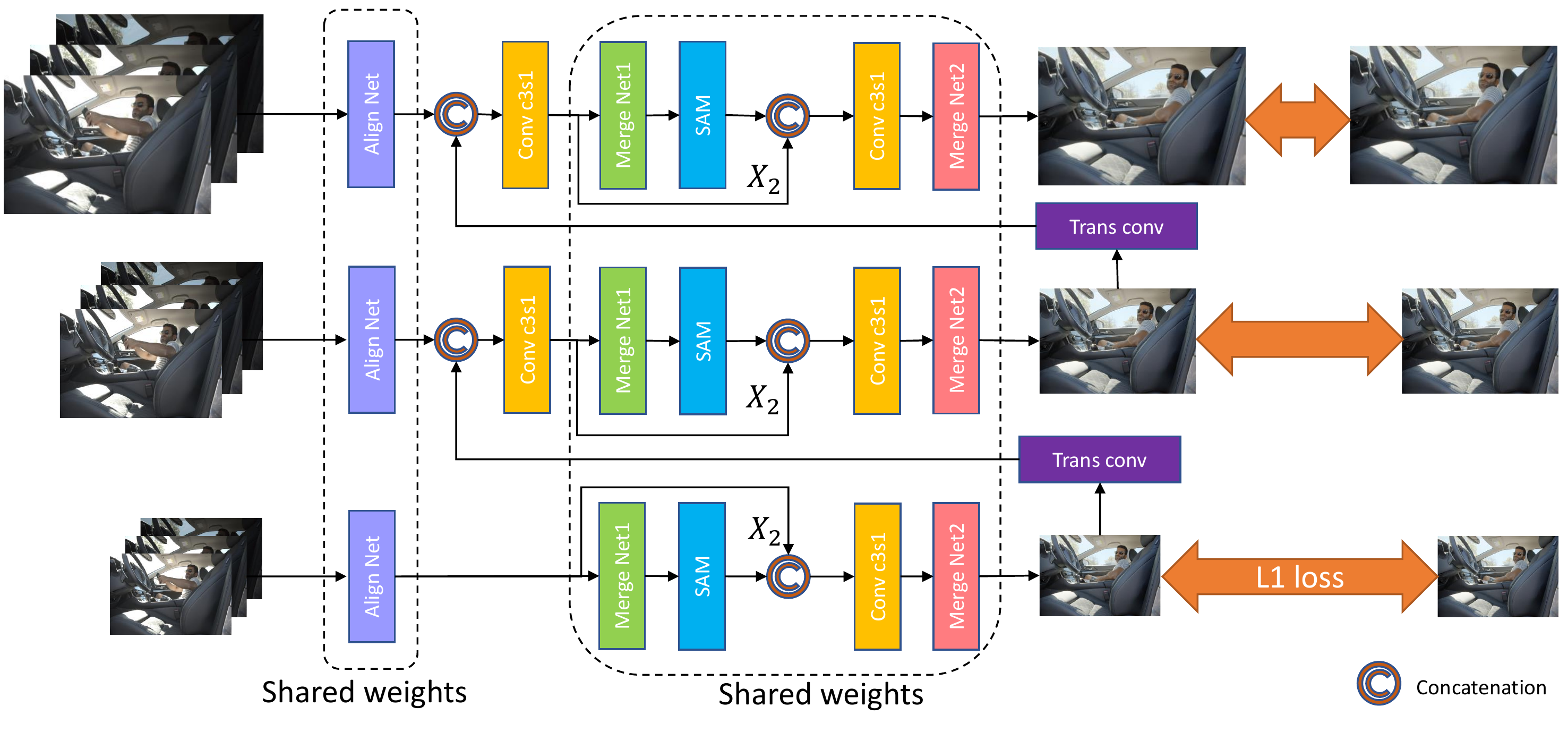}
		\caption{The overview of our framework. There are three scales in our network. Each scale contains alignment and two fusion steps. The two fusion stages are connected by a SAM\cite{zamir2021multi}.}
		\label{fig:overview}
	\end{center}
\end{figure*}

\section{Related work}
In the existing mainstream mobile phone imaging pipelines, dynamic range fusion mainly occurs in the linear high-bit radiance domain or non-linear 8-bit image domain. Therefore, we divide HDR methods into two broad categories: fusion in image domain and radiance domain.

\subsection{Fusion in image domain}
MEF methods\cite{qu2022transmef, zhu2020eemefn} in 8-bit image domain have been proposed over the past few decades. 
These kind of methods differs on weight calculation, weight map smoothing, multi-scale implementation and detail enhancement. 
MEF directly fuses image sequences into one image, which is simple to operate, but has ghosting artifacts.
Specifically, the classic MEF method by \cite{mertens2009exposure} calculate the weight map using contrast, color saturation, and exposure measurements. The fusion is done in a multi-scale framework, where the input image is decomposed into a Laplacian Pyramid and weight map smoothed within Gaussian pyramid. While computationally efficient, this approach suffers detail lost. \cite{li2012detail} enhanced detail results of Mertens solving quadratic optimization problem.  \cite{kou2018edge} replace Gaussian smoothing using gradient domain guided smoothing in further to reduce halo effect. \cite{ancuti2016single} provides a fast single-scale approximate by applying Gaussian filtering to the weight map and adding back using the extracted details second order Laplacian filter. 
\cite{ma2017robust} proposed a structural patch decomposition method that decomposes image patches into three parts: intensity, structure and average intensity. Three patch component are processed  separately and merged finally. 
\cite{li2020fast} further enhances this structural patch decomposition by reducing halos and preserving edges.
When dealing with dynamic scenes, explicit motion pixel detection or alignment is required, such as optical flow \cite{sen2012robust}, image gradients \cite{zhang2010gradient}, SIFT \cite{liu2015dense}, Shannon entropy \cite{jacobs2008automatic}, and structural similarity. Due to the difficulty in detecting moving pixels in the nonlinear domain, these methods are prone to ghosting effects in complex scenes.

\subsection{Fusion in radiance domain}
Previous HDR reconstruction methods \cite{debevec2008recovering} in radiance domain first build a radiation map by recovering CRF , also known as inverse tone mapping and then fuse the radiometric values via weighted sum. However, the calculation of CRF is complex and prone to reconstruction errors \cite{chakrabarti2014modeling}.
Fortunately, as sensor response sensitivity increases, popular camera senor in mobile phones can have easy access to  high-bit raw data without sophisticated CRF recovery. 

In recent years, advances in these methods lie in deep network design \cite{niu2021hdr,wu2018deep,yan2019attention,yan2020deep,yan2021towards,yan2022dual} with the available public training set established by \cite{kalantari2017deep}. They aim to learn to align and fuse demosaiced images in an end-to-end fashion.
In 2017, \cite{kalantari2017deep}  first introduced convolutional neural network (CNN) to predicts irradiance from three low dynamics range (LDR) images with different exposures, and camera and object motion. This method needs the optical flow algorithm to align inputs, which is prone to inaccuracy owing to different exposure levels.
\cite{wu2018deep} removed prior optical alignment and translated HDR imaging into a image translation problem by end-to-end learning. \cite{yan2019attention} leveraged attention mechanism to improve the deghosting performance of \cite{wu2018deep}. 
\cite{niu2021hdr} added Generative adversarial network (GAN) loss to enhance the details, but introduce unusual textures and  color distortion.
In addition, CNN-based methods for single-image HDR have also been studied in inverse tone-mapping \cite{eilertsen2017hdr, endo2017deep, liu2020single, santos2020single}. They rely on deep network to restore missing details in the darkest and saturated areas of tone-mapped images.
Although much progress has been made in deep HDR, it is still important to design lightweight networks to address ghosting and loss of detail in complicated scene.

\section{Method}
Given a series of LDR inputs, $I = \{I_1, I_2, I_3\}$, we first use gamma correction to map them into linear domain for eliminating the domain gap.

\begin{equation}\label{XX}
    H_i = \frac{I_i^\gamma}{t_i}, i=1,2,3
\end{equation}

where $t_i$ and $H_i$ represent exposure time and mapping result in linear domain corresponding to LDR inputs respectively. In this paper, we set $\gamma=2.2$ for a simple linear domain transformation following \cite{kalantari2017deep}. Besides, we treat $I_2$ as the reference frame. After correction, we concatenate LDR inputs and their HDR version together before feeding inputs to the network.

\begin{equation}\label{XX}
    X_i = \{I_i, H_i\}, i=1,2,3
\end{equation}

The target of this paper is to take $X_i$ as inputs and finally generate a detailed HDR prediction with little ghosts. We can use a formula to describe this process.

\begin{equation}\label{XX}
    \hat{H} = f(X_1, X_2, X_3, \theta)
\end{equation}

where $f(\cdot)$ denotes our STHDR network and $\theta$ is the learnable weights in our framework.
\begin{figure}
	\begin{center}
		\includegraphics[scale=0.64]{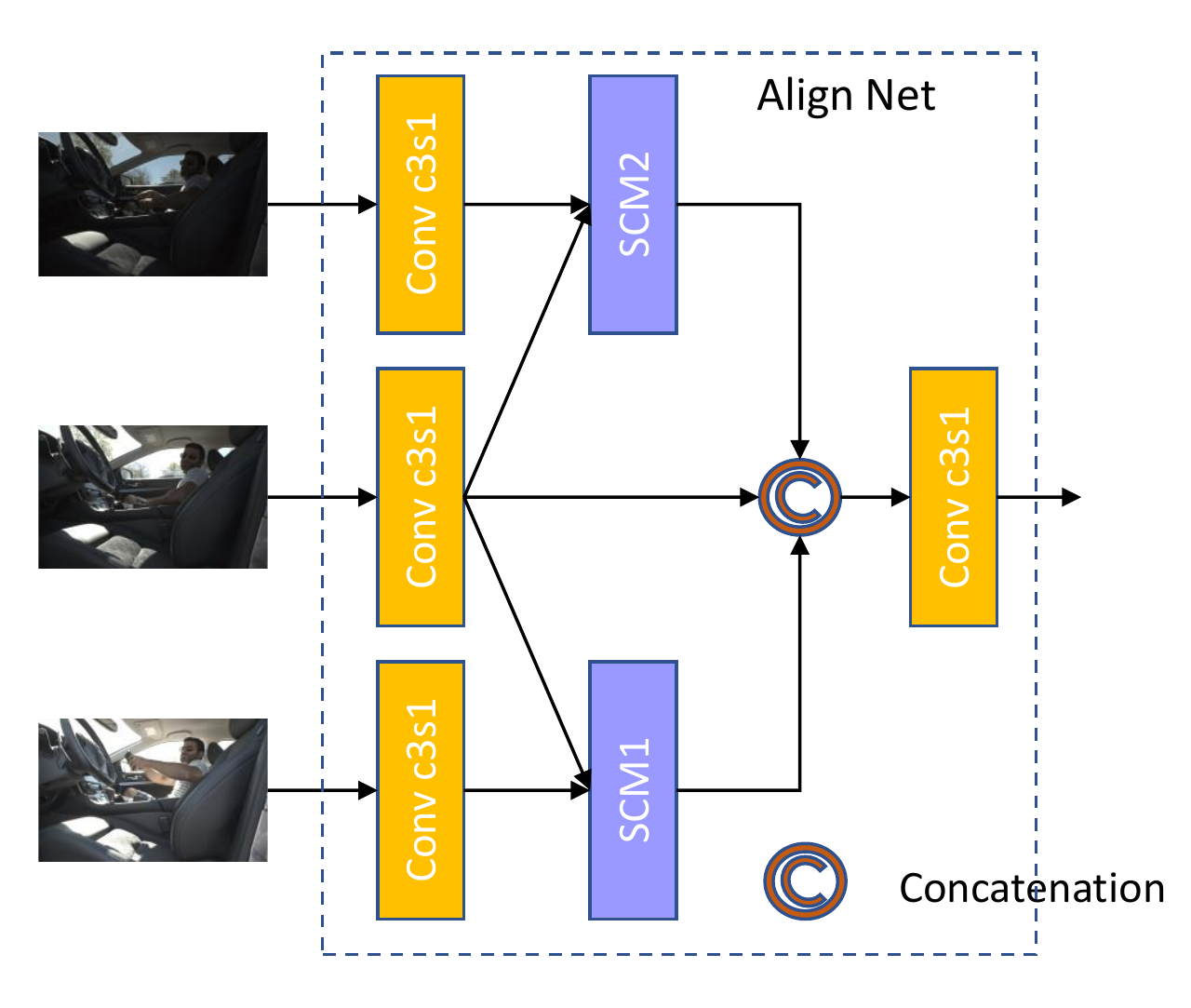}
		\caption{The Align Net in our framework, which contains two SCM and two Convolutions.}
		\label{fig:Alignment}
	\end{center}
\end{figure}

\subsection{Overview}
In this section, we introduce the proposed scale-aware two-stage framework illustrated in Figure.~\ref{fig:overview}. 
Our framework has $n$ scales with three different input sizes. 
For the coarsest $n-th$ scale,  an Align Net is first used to align non-reference frames to the reference in feature space. Then, we concatenate the aligned features with the reference feature together as $Z_0 = \{AF_1, X_2, AF_3\}$, where $AF_i$ are the aligned feature from $SCM_i, i=1,3$. Then, we put the features into two-stage Merge Net for fusion. We employ the supervised attention module (SAM) from \cite{zamir2021multi} to produce an initially predicted HDR output and SAM features in first stage. The initial output has some obvious artifacts, which will be reduced in the second stage.  The SAM features are combined with output of Align Net and fed into another Align Net for finer merging. Finally, we calculate loss using supervisory signal and result of second Align Net.
For other scales, for example $k-th$ layer, $1 \leq k < n$, we concatenate the feature of Align Net and upsampled output of $k+1 -th$ layer. In this way, previous scale results will guide current scale for a better result than a single scale. In this paper, we set $n=3$, which means there are two feature passing across scales. 
\begin{figure}
	\begin{center}
		\includegraphics[scale=0.56]{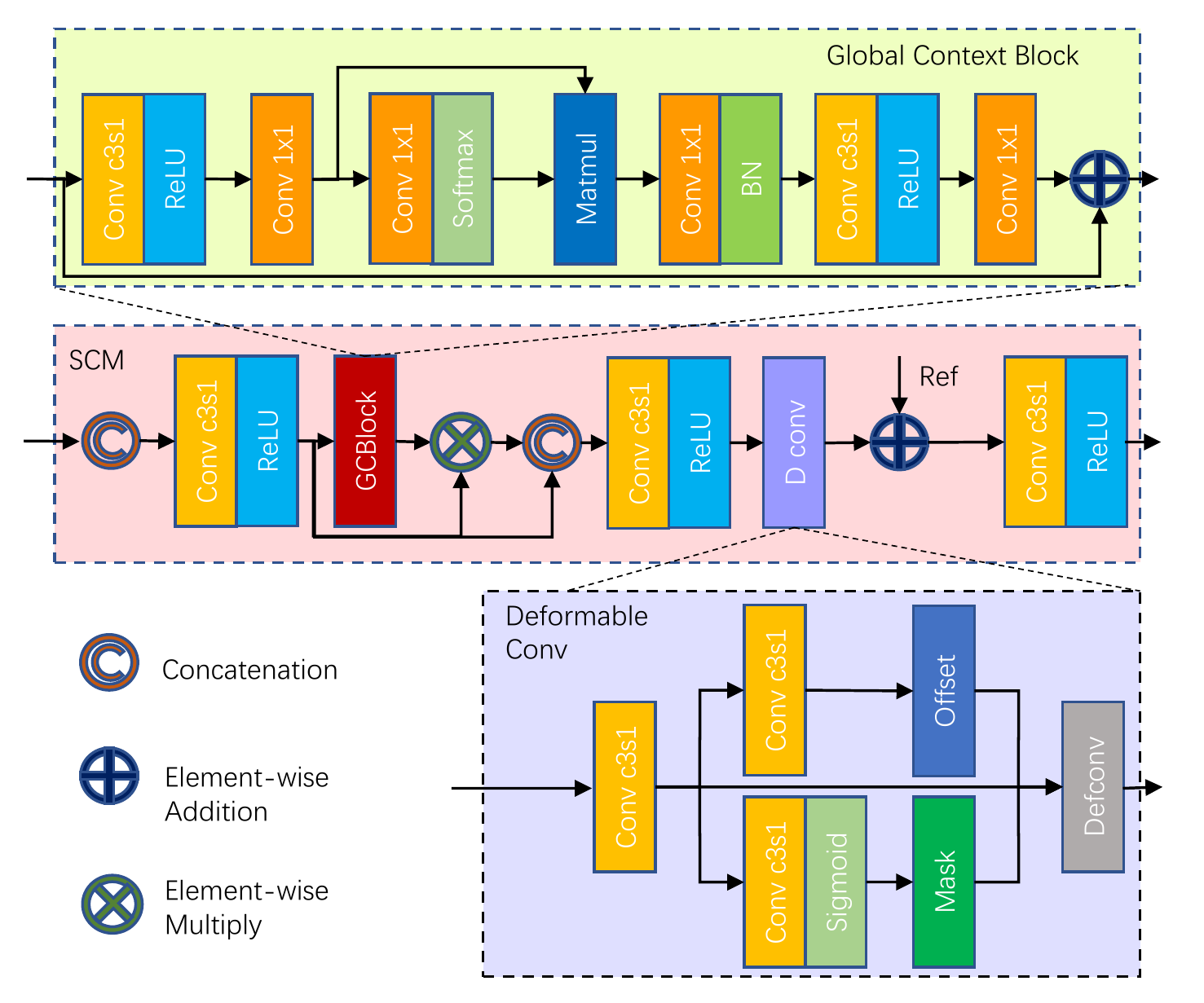}
		\caption{The details of the SCM, which contains a GCBlock and a Deformable Conv. The GCBlock takes account of all information of two inputs. The Deformable Conv find useful details in deformable area. Then add these details into original reference image.}
		\label{fig:SCM}
	\end{center}
\end{figure}
\begin{figure*}
	\begin{center}
		\includegraphics[scale=0.52]{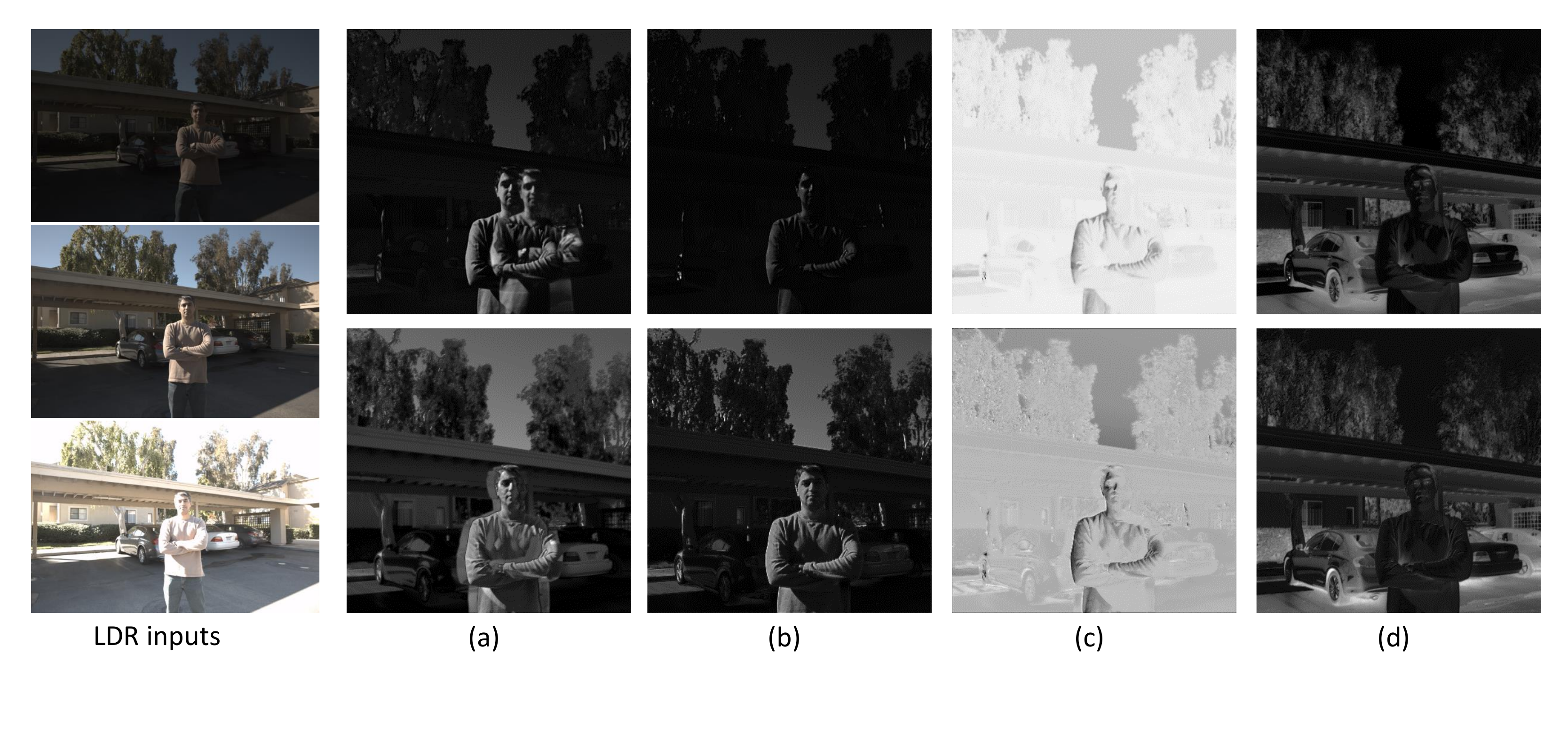}
		\caption{Some middle results of our SCM. (a) is the result of first combination of Conv and LeakyReLU, which has severe ghost shadows. (b) is the compressed result of weighted feature and original one. (c) is the supplementary details produced by deformable conv. (d) is the fianl combination of conv and LeakyReLU in SCM.}
		\label{fig:SCM_results}
	\end{center}
\end{figure*}

\subsection{Align Net}

The target of our Align Net as shown in Figure.~\ref{fig:Alignment} is to align the non-reference feature to the middle reference feature. For achieving this goal, we apply two SCM to align $X_1$ and $X_3$ with $X_2$ respectively, $AF_i = SCM(X_i, X_2), i=1,3$. After we get these well aligned features, we concatenate them together and compress channels for a  cheaper computation cost, $Z = conv({AF_1, X_2, AF_3})$. 

Specifically, in SCM as shown in Figure.~\ref{fig:SCM}, we treat reference image as a template and try to extract useful feature from non-reference images and generate a well aligned feature $AF_i, i=1,3$. First, SCM contains two inputs using a convolution layer: $X=conv(X_i, X_2)$. Then we use a Global Context Block(GCB) \cite{cao2019gcnet} to put global information together which takes into account all of two inputs' values, $W=GCB(X)$. We treat the result $W$ as an importance estimate value for feature of each channel, and use it to choose useful channel like channel attention mechanism, $X_i=X_i \cdot W$, where $\cdot$ means element-wise multiplication. After that, we use a convolution operation to reducing the channel number of $X_i$ and $X_2$, which makes network focus on these useful information,$X_i=conv(\{X_2, X_i\})$. The followed deformable convolution \cite{zhu2019deformable} tries to find corresponding parts between $X_i$ and $X_2$. And the skip connection in the last convolution layer will guide the previous parts of SCM to find the different but useful information, which can help to enhance the details in the final result. The middle results of SCM can be found in Figure.~\ref{fig:SCM_results}. From Figure.~\ref{fig:SCM_results}(a) to Figure.~\ref{fig:SCM_results}(d), we can see that SCM first discards those useless feature maps, then it tries to search meaningful texture and put them to the reference. For different exposures, SCM could find different details which can be observed in Figure.~\ref{fig:SCM_results}(d). 

\subsection{Merge Net}
With the aligned features extracted by Align Net above, we use the encoder-decoder architecture for translation relationship learning.
Inspired by latest advances in image restoration \cite{zamir2021multi, chen2021hinet}, a two-stage network is employed. The output of Stage 1 includes preliminary predicted output and SAM feature. The SAM feature and the features of the reference are concatenated as input of Stage 2, which could further enhance the details and alleviate the artifacts.  

\subsection{Multi-scale Loss Function}

The HDR results predicted by our network are in linear domain, and they need to be reproduced before display in 8-bit image domain. 
Following previous work, for an HDR image $H$, we utilize a $\mu$-law tonemapper to compress the dynamic range  and calculate the loss in tone mapped images. 

\begin{equation}\label{XX}
\tau(H) = \frac{log(1 + \mu H)}{log(1 + \mu)}
\end{equation}

where $\tau$ is the tonemapper. We set $\mu=5000$ and keep $H$ in a range of $[0, 1]$. In our framework, the output contains HDR predictions among all scales. We hope that HDR prediction $\hat{H}$ gets as close to HDR ground truth $H$ as possible in each scale. So we exploit average absolute distance for each gray value.

\begin{equation}\label{XX}
L = \sum_{i=1}^k\lambda_i||\tau(\hat{H}) - \tau(H)||_1
\end{equation}

where $k$ means the number of scale layers, and we set $k=3$ as a simple yet effective implementation. $\lambda_i$ represents the weights for loss in $i$-th layer. 
Considering that each scale can reflect the information of its own frequency band, we set $\lambda_i=1, i=1,2,3$.

\subsection{Implementation}
We implement the proposed framework using Pytorch. The channel numbers of all convolution layers are 32, and kernel sizes are $3 \times 3$ or $1\times 1$. We use LeakyReLU as our activation function. 
We apply Adam optimizer \cite{kingma2014adam} in training. We set the batch size as 16, and initial learning rate as 1e-4 which decreases in a cosine annealing strategy\cite{loshchilov2016sgdr}. We set $\beta$ from 0.9 to 0.999. We set max epoch as 80000, minimum learning rate as 1e-6. For each group of input images, we random crop them to the size of 256 $\times$ 256 for training. After training,  we reduce the batch size to 8, and increase the input size to $384 \times 384$ for fine-tuning. We fine-tuned 2000 epochs. Our model runs on an NVIDIA RTX 3090 GPU.

\section{Experiments}
\subsection{Settings}
We use Kal's HDR dataset\cite{kalantari2017deep} for training and test. This dataset has 74 groups of training data and 15 for testing. There are three LDR images with different exposure bias like {-3, 0, 3} or {-2, 0, 2} and an HDR ground-truth in each group data. To prevent overfitting, we train patches using flip and rotation augmentation.

 \begin{figure}
	\begin{center}
		\includegraphics[scale=0.305]{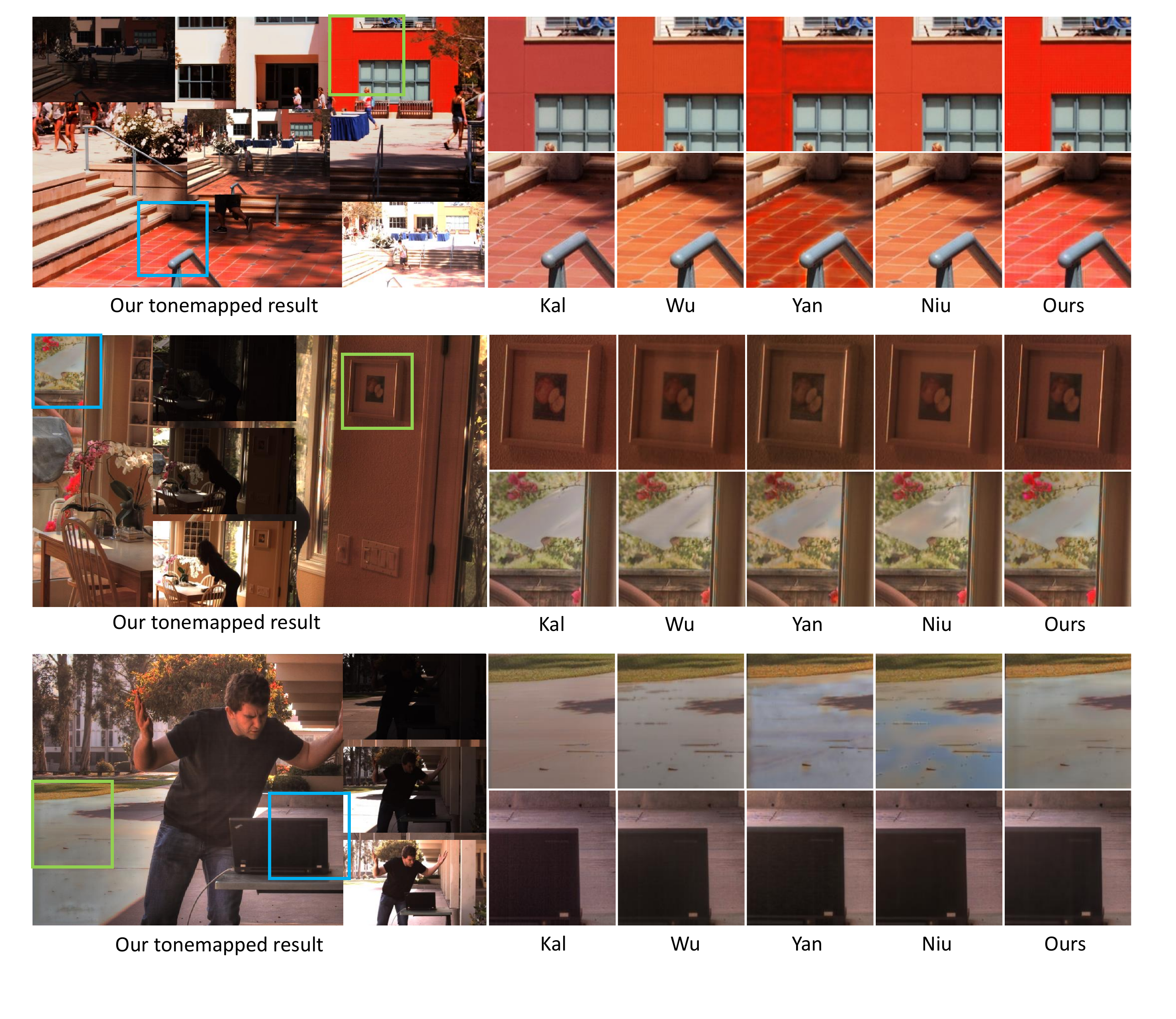}
		\caption{We compare our method in kal's HDR video dataset\cite{kalantari2019deep} with other deep-learning based methods. Our network can also work well in HDR video generation.}
		\label{fig:compare1}
	\end{center}
\end{figure}

 \begin{figure*}
	\begin{center}
		\includegraphics[scale=0.4]{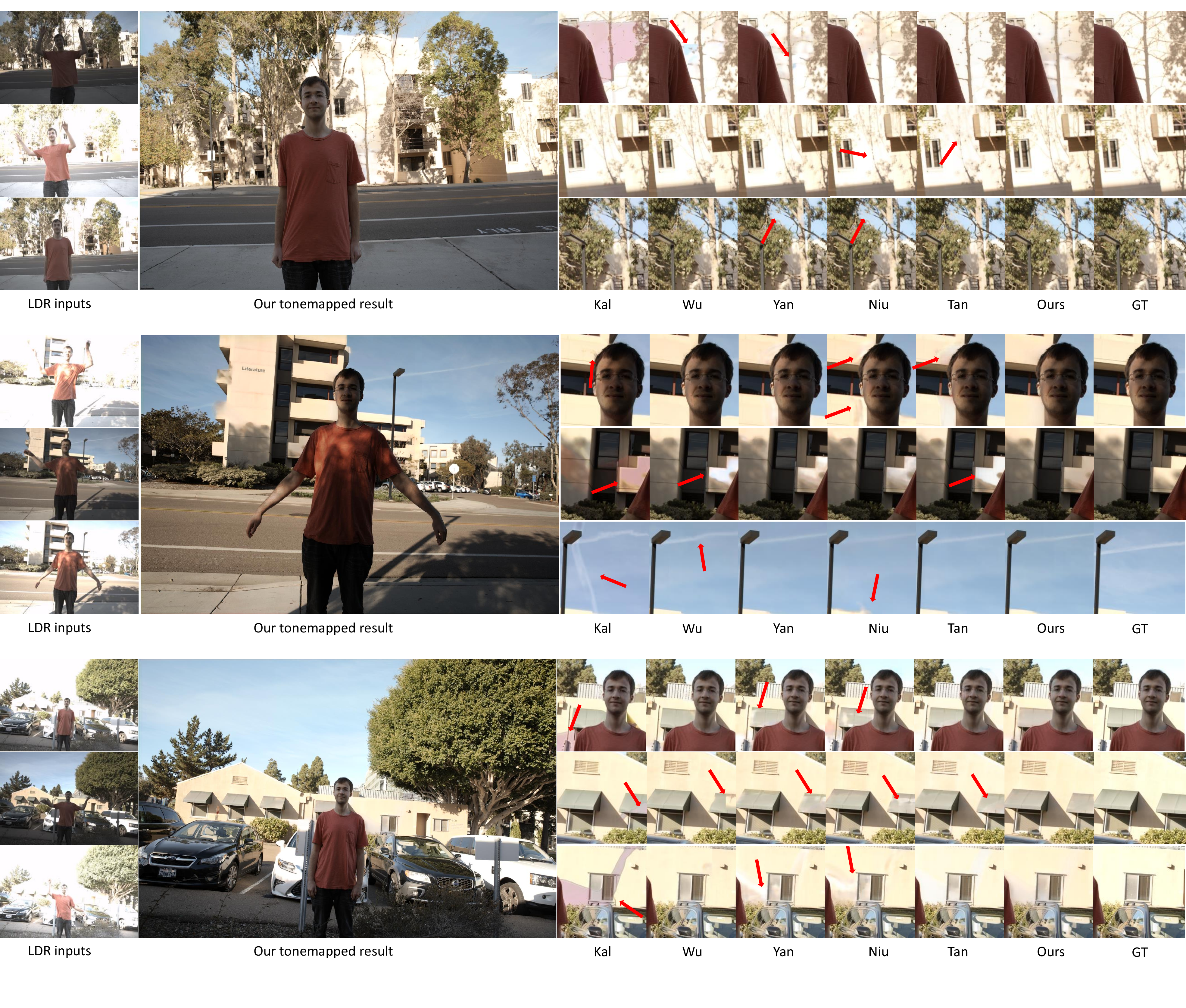}
		\caption{This is the comparison of state-of-arts models on Kal's test data set\cite{kalantari2017deep}. Our method prefers to generate a smoothing result when there are wrong details or occlusion in non-reference frames.}
		\label{fig:image2}
	\end{center}
\end{figure*}

 \begin{figure*}
	\begin{center}
		\includegraphics[scale=0.52]{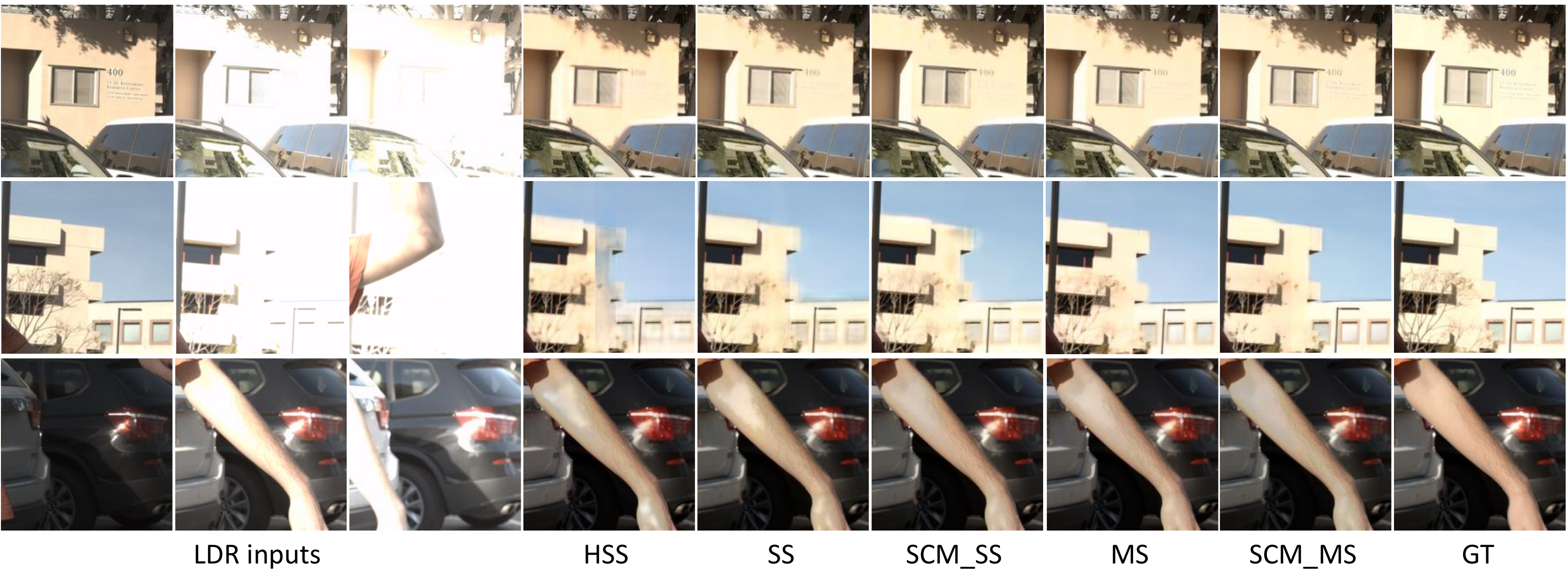}
		\caption{This is the Ablation experiment which shows the characters of strategies and modules.}
		\label{fig:ablation}
	\end{center}
\end{figure*}

\subsection{Comparative experiments}
Since traditional methods are generally worse than deep learning-based methods, 
we compare our network with five other state-of-the-art learning-based methods. Specifically, we choose the direct mode in \cite{kalantari2017deep}, which is a fully convolutional network with gradually decreasing kernel size. The method in \cite{wu2018deep} is a UNet-like architecture which contains several encoders for each LDR inputs. \cite{niu2021hdr} introduces GAN for detail exhibition. \cite{yan2019attention} applies attention mechanism for better deghosting. \cite{tan2021deep} combines super-resolution and HDR imaging tasks (We use the results in normal resolution). For the above methods, we use the officially released code and the given pre-trained model. 

\subsection{Qualitative Comparison}
In Figure.~\ref{fig:image2}, most methods performs well in deghosting, but get poor results in backgrounds. In the contrast, our method could get a better smoothing result in background. Kal's method sometimes could not get a clear result when occlusion or saturation happens in reference frame. Wu's will introduce some holes or highlight regions on background. Our method could get a better and smoothing textures. However, sine our network prefers similar details rather than correct details, it also causes some blur results if the areas in the reference are saturate. We also validate our methods on kal's video database \cite{kalantari2019deep} as shown in Figure.~\ref{fig:compare1}. Our method also could produce competitive results. As we mentioned before, our method also generates smooth results on this dataset when the saturation occurs in the reference image.

\subsection{Quantitative Comparison}
We use PSNR and SSIM \cite{wang2004image} values for each HDR prediction of test images as shown in Table.~\ref{tab:cp}. The name of PSNR-$\mu$ and SSIM-$\mu$ represent the PSNR and SSIM values of tonemapped results using $\mu$-law \cite{kalantari2017deep}. PSNR-L and SSIM-L are results in HDR domain. And we calculate the HDR-VDP-2 \cite{mantiuk2011hdr} after $\mu$-law processing.
We also calculate IL-NIQE \cite{zhang2015feature} scores which do not need reference images as shown in Table.~\ref{tab:cp_non_ref2}. Our method also get comparative scores with these state-of-art methods.

\begin{table}[!htbp]
    
	\centering
	\caption{Quantitative comparisons with different state-of-art methods. All values are the average across results of 15 test data groups}
	\resizebox{\columnwidth}{!}{
	\begin{tabular}{cccccc}
		\hline
		\multicolumn{1}{c}{Method} & PSNR-$\mu$ & PSNR-L & SSIM-$\mu$ & SSIM-L & HDR-VDP-2 \\
		\hline
		\multicolumn{1}{c}{PatchBased} & 41.1074 & 38.8008 & 0.9830 & 0.9749 & 60.6268\\
		\multicolumn{1}{c}{AHDR}& 41.8726 & 41.1964 & 0.9892 & 0.9856 & 62.6731 \\ 
		\multicolumn{1}{c}{HDR-GAN}& 43.9585 & 41.7556 & 0.9906 & 0.9869 & 63.3366 \\ 
		\multicolumn{1}{c}{CNN(Kal)}& 40.1515 & 40.2723 & 0.9807 & 0.9857 & 64.3389 \\ 
		\multicolumn{1}{c}{DHDR}& 42.6703 & 40.8064 & 0.9905 & 0.9882 & 54.3389 \\ 
		\multicolumn{1}{c}{HDR-SR}& 42.7387 & 41.2738 & 0.9887 & 0.9857 & 64.3426 \\ 
		\multicolumn{1}{c}{Ours}& 42.8409 & 40.4269 & 0.9893 & 0.9853 & 62.2958 \\ 
		\hline
		\label{tab:cp}
	\end{tabular}
	}
\end{table}

 \begin{table}[!htbp]
    
	\centering
	\caption{Quantitative comparisons in IL-NIQE with different state-of-art methods.}
	\resizebox{\columnwidth}{!}{
	\begin{tabular}{cccccc}
		\hline
		\multicolumn{1}{c}{Images} & CNN(Kal) & HDRGAN & AHDR & DHDR & Ours \\
		\hline
		\multicolumn{1}{c}{001}& 19.3265 & 20.1267 & 20.3609 & 19.8337 & 19.6031\\
		\multicolumn{1}{c}{002}& 19.1800 & 20.2549 & 19.8961 & 20.1478 & 19.4825 \\ 
		\multicolumn{1}{c}{003}& 21.6377 & 21.6353 & 21.0743 & 21.3740 & 21.6295 \\ 
		\multicolumn{1}{c}{004}& 20.8817 & 22.2027 & 22.4030 & 21.9445 & 22.2576 \\ 
		\multicolumn{1}{c}{005}& 15.6507 & 15.9832 & 16.0110 & 16.0153 & 15.8310 \\ 
		\multicolumn{1}{c}{006}& 17.3841 & 18.4503 & 18.2891 & 18.2964 & 18.2346 \\ 
		\multicolumn{1}{c}{007}& 19.0970 & 20.7697 & 20.4151 & 20.4129 & 20.9783 \\
		\multicolumn{1}{c}{008}& 18.8723 & 18.5262 & 18.6848 & 18.6493 & 18.4129 \\ 
		\multicolumn{1}{c}{009}& 21.5188 & 21.4812 & 21.6147 & 22.2079 & 22.1798 \\ 
		\multicolumn{1}{c}{010}& 18.2286 & 18.2972 & 18.9030 & 18.7812 & 18.8854 \\ 
		\multicolumn{1}{c}{BarbequeDay}& 17.1350 & 17.3294 & 16.9995 & 16.7762 & 17.1796 \\ 
		\multicolumn{1}{c}{LadySitting}& 21.1568 & 21.8552 & 21.6551 & 21.6015 & 21.5995 \\ 
		\multicolumn{1}{c}{ManStanding}& 18.4690 & 19.3215 & 19.4111 & 19.6164 & 19.4073 \\ 
		\multicolumn{1}{c}{PeopleStanding}& 15.7115 & 15.9554 & 15.9943 & 15.9693 & 15.7968 \\ 
		\multicolumn{1}{c}{PeopleTalking}& 20.9976 & 22.2353 & 22.2870 & 22.6154 & 22.5687 \\ 
		\multicolumn{1}{c}{Avg}& 19.0165 & 19.6283 & 19.5999 & 19.6161 & 19.6031 \\ 
		\hline
		\label{tab:cp_non_ref2}
	\end{tabular}
	}
\end{table}
 
\subsection{Ablation Experiment}

We validate the proposed framework and estimate the importance of each module. We explore improvement brought by each component by finishing this ablation experiment. 
\begin{itemize}
    \item \textbf{HSS}. This is the baseline in our experiment, which is a simple UNet with HIBlocks. 
    \item \textbf{SS}. SS means a single-scale two-stage network. 
    \item \textbf{MS}. MS represents a multi-scale two-stage network. It consists of a two-stage merge net and several convolutions for controlling channels. Note that, merge net and some convolution are shared weights across scales.
    \item \textbf{SCM$\_$SS}. This is an SS plus an Align Net.
    \item \textbf{SCM$\_$MS}. This is the complete version of our framework, which contains an Align Net and two-stage Merge Net. The SCMs and Merge Net are shared weights across scales.
\end{itemize}

The ablation study results are shown in Table.~\ref{tab:ab} and Figure.~\ref{fig:ablation}. By comparing HSS and SS, we can find that two-stage strategy brings twice parameters and computation cost. However, this pays off with a significant performance boost. The multi-scale implementation brings a significant improvement to the quality of HDR results at the cost of a small number of parameters and a 1.5$\times$ computational cost, especially in HDR-VDP-2. In the ablation experiment, we find that scale-aware multi-stage strategies not only bring improvement in test scores, but also enhance the detail performance. However, because they lack specialized alignment modules, the saturated areas and occluded regions are not handled well. Our Align Net can effectively solve this problem. It could offer an well-aligned feature with abundant details and without ghosts. 

\begin{table}[!htbp]

	\centering
	\caption{Quantitative comparisons with different models. All values are the average across results of 15 test data groups.}
	\resizebox{\columnwidth}{!}{
		\begin{tabular}{cccccc}
			\hline
			\multicolumn{1}{c}{Method} & PSNR-$\mu$ & PSNR-L & SSIM-$\mu$ & SSIM-L & HDR-VDP-2 \\
			\hline
			\multicolumn{1}{c}{HSS} & 39.3389 & 38.7547 & 0.9809 & 0.9809 & 54.9542 \\
			\multicolumn{1}{c}{SS} & 41.6693 & 39.8216 & 0.9869 & 0.9833 & 59.3218 \\
            \multicolumn{1}{c}{MS} & 42.1188 & 40.2144 & 0.9882 & 0.9844 & 61.7550 \\
            \multicolumn{1}{c}{SCM$\_$SS} & 42.1324 & 40.2872 & 0.9889 & 0.9846 & 60.2071 \\
            \multicolumn{1}{c}{SCM$\_$MS} & 42.8409 & 40.4269 & 0.9893 & 0.9853 & 62.2958 \\
			\hline
			\label{tab:ab}
		\end{tabular}
	}
\end{table}

\begin{table}[!htbp]

	\centering
	\caption{Quantitative comparisons with different models. All values are the average across results of 15 test data groups.}
	\resizebox{\columnwidth}{!}{
		\begin{tabular}{cccc}
			\hline
			\multicolumn{1}{c}{Method} & Parameters(M) & FLOPs(G) & running time(sec)\\
			\hline
			\multicolumn{1}{c}{AHDR}& 1.4412 & 379.7866 & 3.2300\\ 
			\multicolumn{1}{c}{HDRGAN}& 0.6633 & 90.0631 & 3.5399\\ 
			\multicolumn{1}{c}{CNN(Kal)}& 0.3836 & 100.4932 & 3.4692 \\
			\multicolumn{1}{c}{DHDR}& 16.6063 & 254.8668 & 3.5227 \\
			\multicolumn{1}{c}{SCM$\_$MS}& 1.5641 & 48.1668 & 2.3028 \\ 
			\multicolumn{1}{c}{SCM$\_$SS}& 1.5514 & 36.7321 & 2.3006 \\
			\multicolumn{1}{c}{HSS} & 0.6793 & 13.4207 & 2.1166 \\
			\multicolumn{1}{c}{SS} & 1.4195 & 27.2021 & 2.1262 \\
			\multicolumn{1}{c}{MS} & 1.4598 & 37.1876 & 2.1337 \\
			\hline
			\label{tab:pf}
		\end{tabular}
	}
\end{table}

\subsection{Parameters and Speed}

Given that our model needs to take input images with resolution of $2^x \times 2^x$, we set input as $512 \times 512$ for all methods to test running time and FLOPs. Note that, the models of \cite{niu2021hdr, kalantari2017deep, wu2018deep} are implemented in TensorFlow. For getting a clearer result, we implement them in Pytorch which is only used to calculate the parameters and time cost. In our implementation, they attain same numbers of parameter as Tensorflow versions. Although our models contains similar parameters with \cite{yan2019attention}, we do not apply some modules with large computation and get less FLOPs and time cost. Benefiting from our multi-scale multi-stage strategy and modeling ability of HIBlock, we achieve similar performance with less computation.

The number of parameters of our methods is similar to AHDR, but our FLOPs is one seventh of it. Even though our net are not the smallest framework, but the fastest. Our FLOPs is about half of HDRGAN and CNN, one fifth of DHDR. Not only the fast theoretical performance, our actual running speed is also the fastest among the comparison methods.

\section{Conclusion}

In this paper, we propose a scale-aware two-stage deep neural framework for HDR imaging.
The proposed SCM in alignment can sufficiently exploits useful information among different frames by effectively aligning similar features. 
The combined scale-aware technique and two-stage fusion strategy eventually lead to good deghosting and detail-preserving capacity.
The weight-share coding strategy guarantee that our network is computationally efficient. Extensive quantitative and qualitative experiments are conducted to validate the proposed framework can work well in presence of large movements and serious saturation.


\bibliography{aaai22.bib}

\end{document}